\documentclass[10pt,twocolumn,letterpaper]{article}
\usepackage[accsupp]{axessibility} 
\usepackage{tikz}
\usetikzlibrary{positioning}

\usepackage[pagenumbers]{cvpr}      %

\usepackage{etoolbox}

\usepackage{xcolor}

\usepackage{graphicx}
\usepackage{amsmath, amssymb, amsthm}
\usepackage{booktabs}
\usepackage{adjustbox}
\usepackage{makecell}
\usepackage{algorithm,algorithmic}
\usepackage{blindtext}
\usepackage{comment}
\usepackage{cite}
\usepackage{caption}

\usepackage[accsupp]{axessibility} 
\newcommand{\Psymb}{\mathbb{P}}

\usepackage[pagebackref,breaklinks,colorlinks]{hyperref}

\usepackage[capitalize]{cleveref}
\crefname{section}{Sec.}{Secs.}
\Crefname{section}{Section}{Sections}
\Crefname{table}{Table}{Tables}
\crefname{table}{Tab.}{Tabs.}

\def\cvprPaperID{10915} %
\def\confName{CVPR}
\def\confYear{2022}

\DeclareMathOperator{\argmin}{arg\,min}

\newcommand\blfootnote[1]{%
  \begingroup
  \renewcommand\thefootnote{}\footnote{#1}%
  \addtocounter{footnote}{-1}%
  \endgroup
}

\begin{document}

\title{Leveling Down in Computer Vision: \\ Pareto Inefficiencies in Fair Deep Classifiers}

\author{Dominik Zietlow$^{1,2,\dagger, *}$, Michael Lohaus$^{1,3,\dagger}$, Guha Balakrishnan$^{4}$, Matth\"aus Kleindessner$^{1}$, \\Francesco Locatello$^{1}$, Bernhard Sch\"olkopf$^{1, 2}$, Chris Russell$^{1}$\\
\and
\small $^{1}$Amazon Web Services, T\"ubingen, $^{2}$MPI-IS T\"ubingen, Germany,
\small$^{3}$University of T\"ubingen, Germany, $^{4}$ Rice University, USA
}
\maketitle
\begin{abstract}
\blfootnote{$^\dagger$ Work done during an internship at AWS}
\blfootnote{$^*$ Corresponding author \url{zietld@amazon.de}}
Algorithmic fairness is frequently motivated in terms of a trade-off in which overall performance is decreased so as to improve performance on disadvantaged groups where the algorithm would otherwise be less accurate. Contrary to this, we find that applying existing fairness approaches to computer vision improve fairness by degrading the performance of classifiers across all groups (with increased degradation on the best performing groups).

Extending the bias-variance decomposition for classification to fairness, we theoretically explain why the majority of fairness methods designed for low capacity models should not be used in settings involving high-capacity models, a scenario common to computer vision. We corroborate this analysis with extensive experimental support that shows that many of the fairness heuristics used in computer vision also degrade performance on the most disadvantaged groups. Building on these insights, we propose an adaptive augmentation strategy that, uniquely, of all methods tested, improves performance for the disadvantaged groups.
\end{abstract}
\vspace{-5mm}
\section{Introduction}
High-capacity neural classifiers achieve state-of-the-art performance on most computer vision tasks when assessed by overall test set accuracy. However, researchers have begun examining the unfairness of these models. Here, we use `unfairness' to refer to systematic accuracy differences across protected subgroups defined by human-sensitive attributes like gender and race \cite{hardt2016equality, fairmlbook, berk2018}.\footnotemark
These  differences in accuracy can harm certain population groups, and as a result, numerous strategies for training models that match various measures of accuracy across subgroups have been developed~\cite{zafar2017, agarwal_reductions_approach, awasthi2021, goh2016}. Typically, such methods quantify unfairness by comparing accuracy-related rates between various groups, for example the \emph{Difference of Equal Opportunity} (DEO) \cite{hardt2016equality} compares groupwise true positive rates.

Many recent computer vision fairness studies are motivated by a fairness-accuracy trade-off originally targetting  low-capacity models
\cite{calmon2017optimized,zafar2017,agarwal_reductions_approach,martinez2020,diana2021,abernethy2020adaptive,kleindessner_ordinal_regression} 
where high accuracy on better performing (and often larger) groups comes at the cost of lower accuracy on the worse performing (and often smaller) groups. 
In this case, it is possible to increase fairness by reducing the accuracy on the best-performing group and increasing the accuracy on the worst-performing group, see models A and~B in Figure \ref{fig:pareto}.

\begin{figure}[tp]
  \centering
  \includegraphics[scale=0.42]{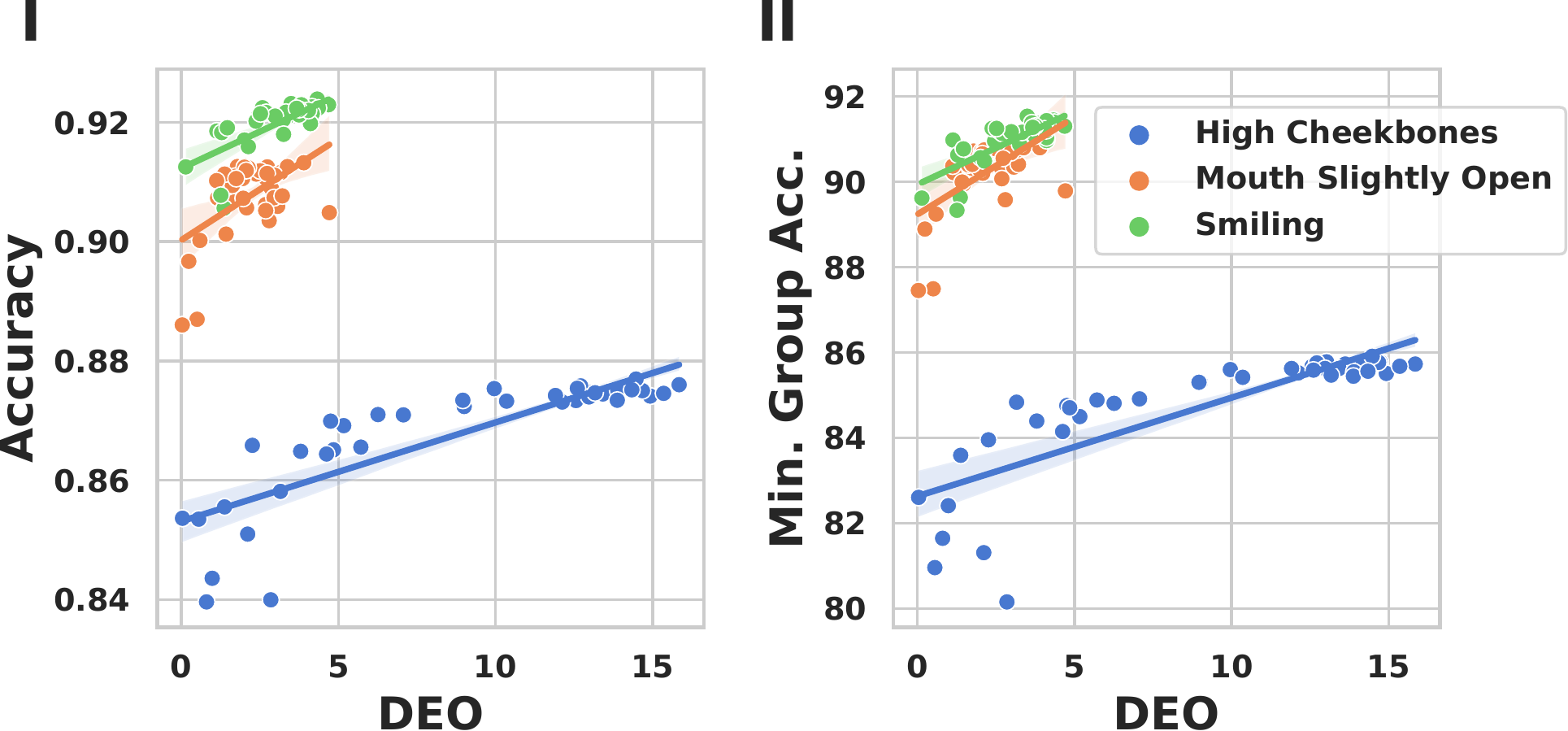}
  \caption{\textbf{Accuracy-fairness trade-off.} We trained multiple fair models on CelebA. The protected attribute is \textit{``Male''} (as annotated in the dataset).%
  Different trade-offs are achieved by varying the strength of a fairness regularizer added to the overall loss~\cite{padala2020fnnc, wick2019}. In~(I), we plot overall accuracy against the fairness measure DEO (low is fair). As expected, improvements in fairness come  with a loss in accuracy. However, this comes at an additional cost; %
  as we see in (II), the worst group accuracy also decreases as DEO improves.\label{fig:celeba_regularizer}}
  \end{figure}

We revisit this trade-off and show that it does not hold when using high-capacity neural classifiers prevalent in computer vision (see Figure \ref{fig:celeba_regularizer}).
Instead, many fairness methods \emph{degrade the accuracy of networks on all groups}, with a greater degradation occurring for the better performing groups.
This increases fairness, but at the cost of producing a worse performing classifier (Figure~\ref{fig:pareto} model~C). 
\footnotetext{In this work, we focus on fairness measures that compare accuracy across groups. These are in contrast to measures such as demographic parity \cite{calders2010three,fta2012}, which matches the proportion of positive decisions per group.}
The phenomenon of balancing fairness by degrading the performance on the better off groups is referred to as \emph{leveling down} in  law and philosophy, where it has received substantial criticism~\cite{holtug1998egalitarianism,doran2001reconsidering,mason2001egalitarianism,
brown2003giving,christiano2008inequality}. The behavior we are concerned with is even more extreme than the typical levelling down -- rather than just lower the performance on the best performing groups, every group is worse off. %

If fairness methods decrease performance for all groups, they are \emph{Pareto Inefficient} with respect to group accuracy and should not be used in contexts where the accuracy of any group is a primary concern.\footnote{Various works in fairness have made use of \emph{Pareto Efficiency}. It has been used both to refer to trading-off global accuracy against notions of fairness\cite{valdivia2021fair}, and for the notion of trading-off per group accuracies against each other\cite{martinez2020}. We only refer to the second case.}
For example, we find that for classifiers on CelebA \cite{liu2015faceattributes} regularized by a DEO fairness measure, the increased fairness comes at the cost of degraded performance for every group including the worst performing group (see Figure \ref{fig:celeba_regularizer}). 
We attribute this problem to two issues:
 \begin{figure}[tp]
    \centering{
  \includegraphics[scale=0.42]{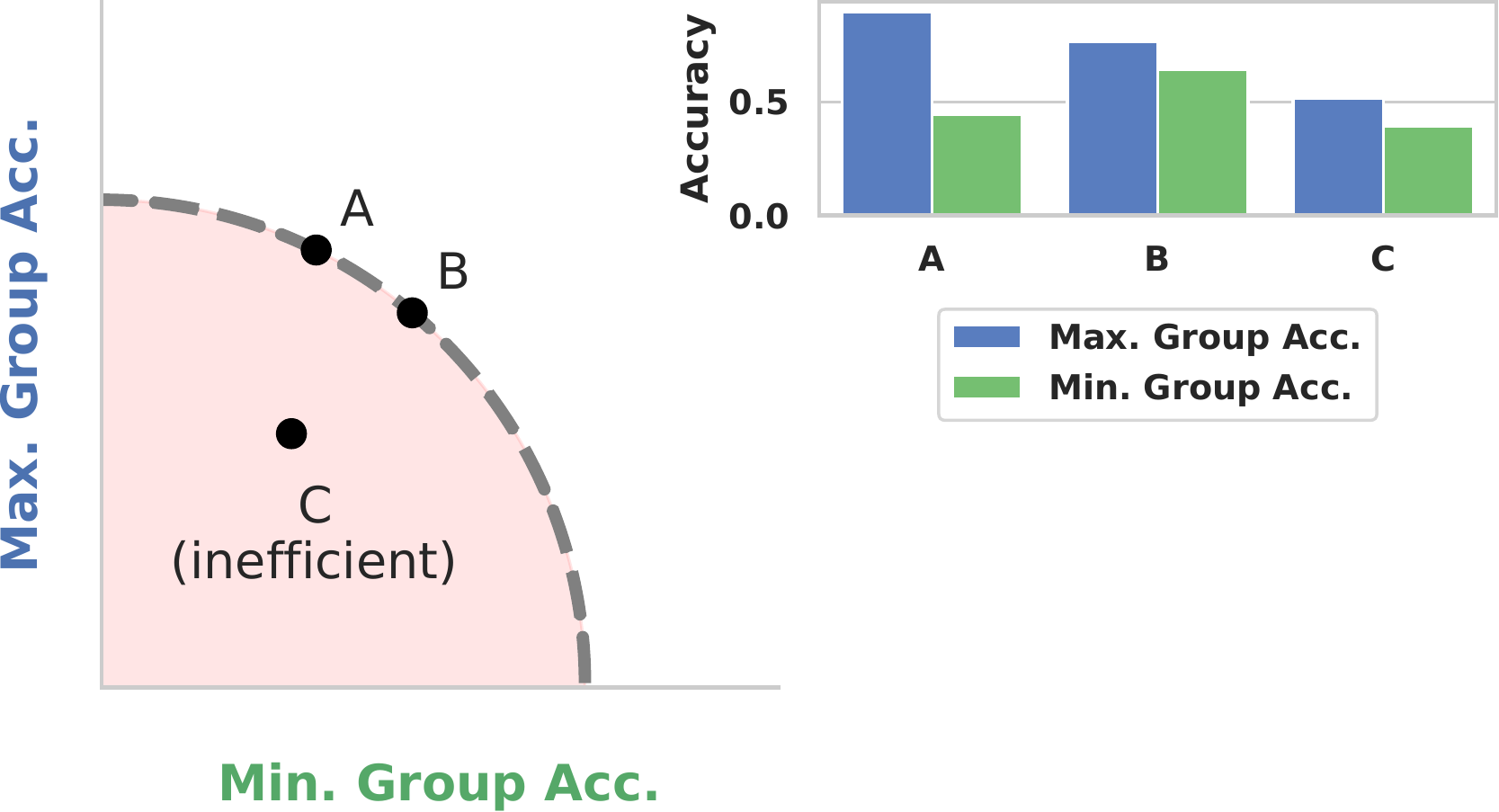}
  \caption{\textbf{Pareto curve.} We depict the typical trade-off assumed by most computer vision fairness studies. The accuracies of a family of classifiers on its best- and worst-performing groups form a \emph{Pareto curve} (dotted gray line). Points A and B are maximally efficient configurations that lie on the curve; B is fairer and has a lower accuracy difference between the groups (see bar plot, right). Point C is as fair as B, but is inefficient because it reduces the accuracies of both groups. Applying accuracy-based fairness methods to deep networks tend to result in inefficient configurations like C.\label{fig:pareto}}}
\end{figure}

\paragraph{High-capacity classifiers fit training data nearly perfectly:} 
Most methods within the fairness community are designed for low-dimensional data where a classifier cannot fit data from multiple distinct distributions well, even on training data \cite{abernethy2020adaptive,agarwal_reductions_approach,calmon2017optimized,diana2021,hardt2016equality,kleindessner_ordinal_regression,martinez2020,zafar2017}. This is not the case in computer vision, where high-dimensional data and high-capacity models mean that near-zero training error is common~\cite{zhang2017understanding}. Fairness notions that are aligned with the perfect classifier are therefore trivially satisfied on training data. \emph{Unfortunately, most existing methods do not take this into account and enforce fairness constraints on the training set \cite{agarwal_reductions_approach,Beutel2017DataDA,diana2021,donahue2016bigan,donini2018,dumoulin2016adversarially,kim2019learning,kleindessner_ordinal_regression,lohaus2020,martinez2020,padala2020fnnc,wang2020fair,zafar2017}}.
\paragraph{Inappropriate evaluation of fairness methods:} 
Most papers presenting fairness methods report a combination of accuracy and a fairness measure such as DEO, and take a decrease in accuracy and an improved fairness measure as an indication that the method is successfully trading off fairness against accuracy \cite{wang2020fair,ramaswamy2020gandebiasing,agarwal_reductions_approach,zafar2017,wang2019balanced,quadrianto2019discovering}.
The crucial question of whether the learned models work better for more disadvantaged groups remains unanswered.
Consequently, the choice of metrics may mask a systematic deterioration of classifiers, where accuracy decreases across all groups, and not just in the high-accuracy  groups.\\
\\~
To address these limitations, we make three contributions:
\begin{enumerate}
    \item We revisit the bias-variance trade-off in a simple \textbf{decomposition of fairness on test data into training error (bias) and generalization error (variance)}, and observe that in situations where the training error goes to zero, any measure of fairness must be dominated by generalization error. As such, methods not using held-out data for fairness constraints cannot work for the high-capacity classifiers common to computer vision.
    \item We perform an extensive \textbf{evaluation of existing fairness methods} and show that their reported improvement in fairness metrics is accompanied by \textbf{worse performance across all groups}.
    \item To confirm our theoretical analysis that \textbf{better generalization is key} to improving performance on the most disadvantaged groups, we explore the use of \textbf{data augmentation combined with adaptive sampling}.
    We propose a novel GAN-based augmentation and show that it improves fairness by improving accuracy on the most disadvantaged groups on the CelebA dataset. 
\end{enumerate}

\section{Related Work}
\paragraph*{Notions of fairness:}
We focus on notions of fairness that aim to balance classifier errors across population subgroups, for example, by matching error rates across genders or different racial groups.
See \cite{Verma2018} for a summary of other forms of fairness, and \cite{wachter2021fairness} for when such definitions are inappropriate. 
One common fairness measure is `equal opportunity' (EO) \cite{hardt2016equality}. This requires a classifier to have equal true positive rates (TPR) on each subgroup. We use $\cal A$ to refer to the set of protected attributes, such as race or gender, $Y\in\{0,1\}$ to refer to the ground-truth label of a datapoint, and $\hat{Y}\in\{0,1\}$ the prediction of $Y$. For two groups $a, a^\prime \in \mathcal{A}$ the violation of equal opportunity is measured by the \emph{difference in equal opportunity} (DEO) defined as 
\begin{align}
\medmuskip=2mu
\thinmuskip=2mu
\thickmuskip=2mu
\begin{split}
\lvert &\Psymb(\hat{Y}=1|Y=1, A=a) - %
\Psymb(\hat{Y}=1|Y=1, A=a') \rvert.
\end{split}
\label{eq:def_deo}
\end{align}
We measure the \emph{difference in equalized odds} (DEOdds) by
\begin{align}
\medmuskip=2mu
\thinmuskip=2mu
\thickmuskip=2mu
\begin{split}
\sum_{y} \lvert &\Psymb(\hat{Y}=1|Y=y, A=a) - %
\Psymb(\hat{Y}=1|Y=y, A=a') \rvert.
\end{split}
\label{eq:def_deodds}
\end{align}

Another fairness notion of recent interest is min-max fairness \cite{minmax2020,diana2021}. In contrast to EO (and the other group fairness notions) min-max fairness does not equalize a statistic across groups, but solely strives to decrease the classification error for the subgroup with the highest error as much as possible. Formally, min-max fairness methods optimize
\begin{align}\label{eq:def_minmax_fairness}
    \min_{\hat{Y}} \max_{a\in\mathcal{A}} \Psymb(\hat{Y}\neq Y | A = a).
\end{align}
As a measure of min-max fairness we report the minimum group accuracy (\ie, $1-\text{error}$) and similarly the minimum group TPR.
Note that the minimum group TPR is trivially maximized by the constant classifier~$\hat{Y}=1$.

While many variants of these error-matching fairness definitions exist\footnote{As a  starting point, \cite{wachter2021fairness} identified 17 accuracy-based measures excluding min-max fairness.}, the motivation for all measures is broadly the same: to prevent a tyranny of the majority, in which accuracy on smaller groups of people are neglected to achieve higher accuracy on larger groups.       
\vspace{-0.8em}
\paragraph*{Fairness in computer vision:}
Following foundational work\cite{gendershades} that identified the systematic failings of face analysis systems on particular racial and gender demographics, algorithmic fairness has attracted growing attention in the computer vision community. Image datasets are known to be biased due to sampling inequalities~\cite{albiero2020analysis,ponce2006dataset, torralba2011unbiased}, and human face datasets have been particularly scrutinized~\cite{balakrishnan2021towards,klare2012face, kortylewski2019analyzing,kortylewski2018empirically,merler2019diversity} as models trained on these data can exhibit systematic failings with respect to attributes protected by the law~\cite{kleinberg2018discrimination}. Multiple approaches to mitigating dataset bias include collecting more diverse examples~\cite{merler2019diversity}, using image synthesis to compensate for distribution gaps~\cite{balakrishnan2021towards, kortylewski2019analyzing, sattigeri2019fairness, singh2021matched, wang2020fair}, and resampling~\cite{li2019repair}.
More recently, approaches have been proposed to mitigate the systematic biases of deep learning-based vision models~\cite{choi2020fair,ganin2015unsupervised, grover2019bias, hendricks2018women,kim2019learning, li2019repair,quadrianto2019discovering, ramaswamy2020gandebiasing,wang2019balanced,wang2019racial}. 

Outside of computer vision, the most common approaches add additional fairness measures to loss functions~\cite{beutel2019fairness,wick2019,padala2020fnnc,lohaus2020, kleindessner_ordinal_regression, risser2020tackling}, enforcing fair representations that are independent of protected attributes~\cite{zemel2013, Beutel2017DataDA, edwards2016, madras2018}, and augmenting training data to promote balance~\cite{ramaswamy2020gandebiasing}. Most of these studies assume an accuracy-fairness tradeoff which we show can lead to suboptimal training or misleading evaluation for deep neural networks.
\vspace{-0.8em}
\paragraph*{Active sampling for min-max fairness:} While the methods of \cite{minmax2020,diana2021} achieve min-max fairness by means of reweighting and retraining, the recent method of \cite{abernethy2020adaptive} uses adaptive sampling and standard SGD updates to minimize a differentiable proxy of \eqref{eq:def_minmax_fairness}. 
This makes the latter method easily applicable to deep neural network training.%

\section{On Accuracy-based Fairness in Low- and High-capacity Classifiers}
\label{analysis}
This section outlines the challenges peculiar to enforcing fairness in the high-capacity classifiers common to computer vision. 
In particular, we focus on accuracy-based notions of fairness that aim to match accuracy across groups. 

As noted by Wachter et al.\cite{wachter2021fairness}, any accuracy-based fairness measure\footnote{Wachter et al. refer to these accuracy-based measures as bias-preserving fairness metrics.}  is trivially satisfied by a classifier with zero error.  However, as the typical datasets (see \cite{Kamiran2012,zafar2017,agarwal_reductions_approach} for a range of examples) used in the  fairness literature are low-dimensional, and with large amounts of label volatility (see Section \ref{bias_fairness:sec}), classifiers with zero error do not occur in practice, even on the training data. 
Training a low-capacity classifier with accuracy-based fairness constraints on the training set is a common approach. Where training error remains high, such approaches can remain effective even if fairness constraints are only enforced over the training data.

This is not the case when enforcing fairness of deep-learning based classifiers on computer vision datasets. Such datasets are empirically \emph{shatterable} \cite{vapnik2013nature}. Even if the images in the training set are randomly relabelled, it is possible to learn a classifier with zero error on the training set \cite{zhang2017understanding}. In such scenarios, accuracy-based fairness definitions are trivially satisfied over the training set.

In the following subsection, we formalize the gap between fairness on training and held-out data and discuss its implications for computer vision.

\subsection{Bias-variance decomposition for classification}
\label{bias_fairness:sec}
It is common in statistics to decompose the error into three terms: irreducible label noise $N$, a bias $B$ representing how well the regressor can fit the dataset, and a variance $V$ representing additional error induced when generalizing to new data.  
The standard decomposition was formalized around the squared loss \cite{10.1162/neco.1992.4.1.1}, extended to the zero-one loss \cite{Kohavi1996BiasPV} and generalized to arbitrary losses \cite{domingos2000unified}. 

We build on the latter formulation, which we summarize in a condensed manner---see \cite{domingos2000unified} for  more details. We consider the task of learning $f\in \mathcal{ F}: \mathcal{X} \rightarrow \mathcal{Y}$, where $f(x)$ should be a good prediction of the label~$y$ of  input point $x$. The quality of a prediction is measured by a loss function $L: \mathcal{Y}\times \mathcal{Y} \rightarrow \mathbb{R}$. The optimal prediction for input~$x$ is 
  $$y_{*}(x) = \argmin_{y'} \mathbb{E}_{y|x} \left[ L(y, y') \right],$$ 
and the expected loss of model $f$ on $x$ is
    $\mathbb{E}_{y|x} \left[ L(y, f(x)) \right]$.
The conditional expectation corresponds to the fact that, in general, the label $y$ is a non-deterministic function of input~$x$. We learn the model~$f$ based on a training set $D_n = \{ (x_1, y_1), \dots, (x_n, y_n) \}$, and in order to remove the dependency on $D_n$ we consider $\mathbb{E}_{D_n,y|x} \left[ L(y, f(x)) \right]$ as the expected loss of $f$ on $x$. The latter is the quantity of interest that we want to decompose into bias, variance and noise.

We define the main prediction $y_{m}(x)$ on $x$ as
\begin{align*}
    y_m(x)=\argmin_{y'} \mathbb{E}_{D_n} \left[ L(f(x), y') \right].
\end{align*}
This allows for definition of the intrinsic noise $N(x)$ (expected error induced by label disagreement for a particular datapoint), the bias $B(x)$ (error induced by systematic model imperfection), and the variance $V(x)$ (error difference from the main prediction):
\begin{align}
    B(x) &= L(y_{*}(x),y_m(x)) \label{eq:bias},\\
    V(x) &= \mathbb{E}_{D_n} \left[ L(y_{m}(x),f(x))\right],\\
    N(x) &= \mathbb{E}_{y|x} \left[ L(y,y_{*}(x)) \right].\label{eq:noise}
\end{align}
As \cite{domingos2000unified} shows, for certain loss functions~$L$ (including but not limited to squared, zero-one loss, and false negative rate\footnote{The use of zero-one loss allows us to consider the bias-variance decomposition of DEOdds, while false negative rate allows us to consider DEO.}) 
we can decompose 
\begin{align}
\text{err}_{x} :=&~\mathbb{E}_{{D_n},y|x} [L(y,f(x))] \\
=&~c_{1}(x) N(x)+B(x)+c_{2}(x) V(x)\label{eq:decomposition}
\end{align}
for some $c_1(x),c_2(x)\in\mathbb{R}$.
\subsection{Expected fairness violations}
We make the simplifying assumption that we want to enforce error parity  across only two groups $A$ and $B$. 
If we define notation for both groups, we use $G$  in place of $A$ or~$B$. We depart slightly from \cite{domingos2000unified} as we consider the per-group error of a classifier trained over both groups rather than the classifier's overall error.
The expected fairness violation for the two-groups case $E_\mathrm{fair}$ can be defined as
\begin{equation}
    E_\mathrm{fair} = |\mathbb{E}_{x \in A} \left[ \text{err}_x \right] - \mathbb{E}_{x \in B} \left[ \text{err}_x \right]|.
\end{equation}
With auxiliary definitions
\begin{align}
    B_G &= \mathbb{E}_{x\in G} [B(X)],\\
    V_G &= \mathbb{E}_{x\in G} [c_{2}(x) V(X)],\\
    N_G &= \mathbb{E}_{x\in G} [c_{1}(x) N(X)],
\end{align}
the fairness violation can be rephrased as
\begin{equation}
    E_\mathrm{fair} = | N_A + B_A + V_A - (N_B + B_B + V_B)|.
\end{equation}

\subsubsection{Fairness for low-capacity classifiers} 
For low-capacity models, the variances are strongly dominated by the noises and biases \cite{hastie2009}. 
That is $N_G + B_G \gg V_G$ and the fairness violation can be approximated as
\begin{equation}
    E_\mathrm{fair} \approx | N_A + B_A - N_B - B_B |.
\end{equation}
Importantly, $N_G + B_G$ can be directly estimated from the per group training set, and as such methods that match the loss across different groups at training time are likely to work here. Indeed, many methods \cite{hardt2016equality,zafar2017,agarwal_reductions_approach} enforce fairness by explicitly balancing errors on the training set,  under the reasonable assumption that these fairness constraints will generalize well to unseen data.
The work\cite{donini2018} addressed this by minimizing a combination of unfairness on the training set and an upper bound of the generalization error.   However, in practice, the majority of approaches are effective without considering generalization error.

\subsubsection{Fairness for high-capacity classifiers} In contrast, for classification tasks in computer vision, the typical behavior is very different. While human labelling of image data is often a noisy process\cite{northcutt21}, the majority of computer vision datasets do not model label noise and often either only collect one label per datapoint\cite{liu2015faceattributes}, or explicitly denoise the collected labels\cite{deng2009imagenet} and only use the most common label per datapoint as the true datapoint.\footnote{\ie, given an image labeled as cat twice, and dog once, the image is simply treated as being labelled cat.} As such, when we treat dataset acquisition as a stochastic process that can only assign one label per datapoint, eq.~\eqref{eq:noise}, which measures the expected loss caused by disagreement between labels, is zero. This point is key to our argument; where multiple disagreeing labels are collected for individual datapoints, it may be possible for methods to improve these forms of fairness without making use of held-out data. 

For a computer vision classifier trained to convergence on a fixed dataset, the bias terms also disappear \cite{yang2020}.  
Here, data is high-dimensional and models have essentially arbitrarily high capacity. Hence, the perfect classifier, which has zero error and predicts the ground-truth label for each datapoint, lies in $\mathcal{F}$. Hence, $V_G \gg B_G\approx 0$, and the fairness violation is dominated by the generalization error 
\begin{equation}
    E_\mathrm{fair} \approx |V_A - V_B|, \label{eq:gen-error}
\end{equation}
\ie, the fairness violation is predominantly determined by the difference of the variances.
Others have made related observations, for example, \cite{wang2020fair} emphasized the importance of data augmentation when reweighting samples, to prevent the gradients going to zero. However, even with augmentations, error on the training set remains much lower than the error on held-out data, and the variance  still dominates.

Given this%
, it is unsurprising that these methods for enforcing fairness on a training set are not popular in computer vision, and instead research typically centers around heuristics, such as reweighting samples \cite{li2019repair}, training multiple classifiers on data subsets, and averaging them  \cite{saerensAdjusting2002,Royer2015ClassifierAA}, or data augmentation \cite{kortylewski2019analyzing}.  With such a wide range of approaches, it is not possible to formally analyze them, beyond saying that they do not use  held-out validation data to estimate generalization error, but they all successfully trade-off accuracy against a wide range of fairness metrics.

\subsubsection{Rethinking fairness measures in computer vision}
\label{sec:rethinking}
We have shown it is not possible to predict per-group error rates on the test set without computing fairness measures on held-out data. Yet the vast majority of methods do not do so, and as demonstrated in Section~\ref{sec:experiments}, still consistently improve fairness according to standard metrics such as equalized odds or equal opportunity. How is this possible?

One  answer lies in the definitions of equalized opportunity~\eqref{eq:def_deo} and odds~\eqref{eq:def_deodds} -- they are perfectly satisfied by random or constant classifiers. As such, it is possible to increase fairness by decreasing performance on all groups, rather than by rebalancing the errors from one group to another. 

Empirically, we find that this is exactly what occurs in practice: improvements to DEO and DEOdds are accompanied by a degradation in the accuracy for all groups, and not a rebalancing of accuracy from one group to another (see Figures~\ref{fig:celeba_regularizer},~\ref{fig:performance_overview}). 
While it is not possible to predict the per-group error for unseen data without making use of held-out data, it is always possible to degrade performance by injecting noise into gradients (which may be happening with regularized approaches \cite{padala2020fnnc, wick2019}), augmenting datasets with inappropriate synthetic examples (which might be happening with \cite{ramaswamy2020gandebiasing}), or by a wide range of well-motivated and sensible-sounding heuristics, including early stopping.  

In light of this issue, when evaluating accuracy-based model fairness, \emph{we recommend choosing a fairness metric that explicitly requires improving performance on the disadvantaged groups.} In our experiments, we measure the accuracy of the worst performing group, \ie, min-max fairness of eq.~\eqref{eq:def_minmax_fairness}.
Another natural choice would be the true positive rate on the worst performing group, but this is easily gamed by classifiers with high false positive rates.

\section{Improving accuracy on disadvantaged groups with synthetic data}
In the previous section, we presented the importance of evaluating fairness based on the  worst-performing group's accuracy. Next, we consider how to use this insight, along with the result in~\eqref{eq:gen-error} to better incorporate fairness during the model \emph{training}. Expression~\eqref{eq:gen-error} shows that fairness violations in high-capacity models are dominated by the variance  of the bias-variance decomposition. 
Consequently, one way to improve fairness without decreasing the performance for all groups is to decrease the variance of the worst-off group. It is well-known, and experimentally verified \cite{brain1999effect}, that variance typically decreases as the training set increases in size and diversity. Given a fixed dataset, data augmentation is the only way to achieve that.
Taking these insights into account, we propose an adaptive augmentation strategy that increases the diversity of samples for the worst-performing group in a classification task.

There are three technical challenges that we must address with our method. How to:
(i) decide which group(s) requires augmentation 
(ii) generate high fidelity in distribution data, and 
(iii) reliably condition the augmented data on the protected groups and automatically provide target labels. 
We tackle these challenges by: (i) deploying adaptive sampling strategies using held-out data to prioritize worst-performing groups, (ii) using invertible GAN architectures and latent space traversals to edit images, and (iii) proposing g-SMOTE, a generalization of the synthetic minority oversampling technique (SMOTE)~\cite{chawla2002smote}, which produces new labeled training images by traversing GAN latent space.

\subsection{Adaptive sampling}
At each training iteration, we sample a random batch from one of two datasets: the original training dataset, and an extended training dataset. The extended training set is equal to the original training set at initialization. After each iteration, we determine the worst-performing group using a held-out evaluation dataset, augment a random batch from this group (using g-SMOTE in next section), and add it to the extended set. This approach is inspired by past studies~\cite{Richter_2016_ECCV, ravuri2019seeing} that evaluate the generalization when training on partially augmented data.

We select a batch from the original dataset with probability $\lambda$ (and $1-\lambda$ from the extended dataset). The parameter~$\lambda$ allows us to maintain a certain proportion of the original data in the training set, preventing the extended training set from becoming predominantly augmented data. Note that this is more involved than simply balancing the number of elements in each group. Differences in generalization performance can stem from strongly imbalanced group sizes, but also from characteristics of either group that make generalization intrinsically harder.
\begin{algorithm}
\caption{Adaptive Sampling}
\begin{algorithmic}[1]
  \scriptsize
  \STATE \textbf{Inputs:} 
  
  Hyper-parameter $\lambda \in [0,1]$
  
  Train dataset $D_\text{Train} = \{ (x_0, y_0), (x_1, y_1), \dots \}$, $x_i \in \mathcal{X}$, $y_i \in \mathcal{Y}$
  
  Evaluation dataset $D_\text{Eval} = \{ (x^e_0, y^e_0), (x^e_1, y^e_1), \dots \}$, $x^e_i \in \mathcal{X}$, $y^e_i \in \mathcal{Y}$
  
  Classifier $c_\phi: \mathcal{X} \rightarrow \mathcal{Y}$ (parameterized by $\phi$)
  \STATE \textbf{Initialize:} $D_\text{Aug} := D_\text{Train}$
  \FOR{$i = 1, \dots, n_\text{training steps}$}
  \STATE With probability $\lambda$, uniformly sample $(x_i, y_i) \in D_\text{Train}$, otherwise sample $(x_i, y_i) \in D_\text{Aug}$
  \STATE Update $\phi$ according to learning objective
  \STATE Determine weakest group based on learning objective and $D_\text{Eval}$ and augment corresponding $x_\text{Aug}, y_\text{Aug}$ from that group
  \STATE $D_\text{Aug} \leftarrow D_\text{Aug} \cup \{(x_\text{Aug}, y_\text{Aug})\}$
  \ENDFOR
\end{algorithmic}
\label{alg:adaptive_sampling}
\end{algorithm}\vspace{-0.5em}
\subsection{Generalized SMOTE: g-SMOTE} 
Given images from the original dataset, we need a procedure to generative new synthetic images along with attribute labels. To accomplish this, we combine SMOTE~\cite{chawla2002smote} with the rich representational and generative capabilities of modern GANs. SMOTE is a simple sampling strategy to overcome imbalanced data that is based on linear feature interpolation between a datapoint and a random datapoint within the set of its $k$ nearest neighbors. It has inspired many other approaches \cite{10.1007/978-3-540-39804-2_12, 4633969, 10.1007/11538059_91, zhang2018mixup}.
Modern invertible GAN architectures \cite{dumoulin2016adversarially, donahue2016bigan, pidhorskyi2020adversarial, ghosh2021invgan} allow one to `embed' images into their latent spaces, which are known to be particularly effective domains to compare and edit images~\cite{balakrishnan2021towards, shen2020interpreting}. We propose to use SMOTE in the GAN latent space to generate new diverse synthetic images along with attribute labels.

We propose a generalized variant of SMOTE (g-SMOTE), that extends the classical SMOTE strategy -- linear interpolation between a datapoint and a random point among the datapoints $m$ nearest neighbors -- to uniform sampling within a $k$-dimensional simplex formed by $k$ of the $m$ nearest neighbors, to improve data diversity. Given a datapoint and its $m$ nearest neighbors with the same target attribute, we randomly choose $k$ neighbors that span a simplex of dimension $k-1$ or smaller in the much higher dimensional GAN latent space.
Within this simplex, we uniformly sample points in latent space and use the GAN generator to render their images. Our key assumption is that the simplex covers a label-consistent volume in latent space, \ie,  that every image stemming from that region shares the same target label.
Algorithm \ref{alg:augmentation} describes the data augmentation procedure, Figure \ref{fig:smote_simplex_sampling} illustrates the interpolation mechanism, and Figure~\ref{fig:g-smote_examples} shows a real-world example for $k=3$.
See supplementary materials for the effects of varying $k$ and a comparison of SMOTE and g-SMOTE.

\begin{figure}
    \centering
    \includegraphics[width=\linewidth]{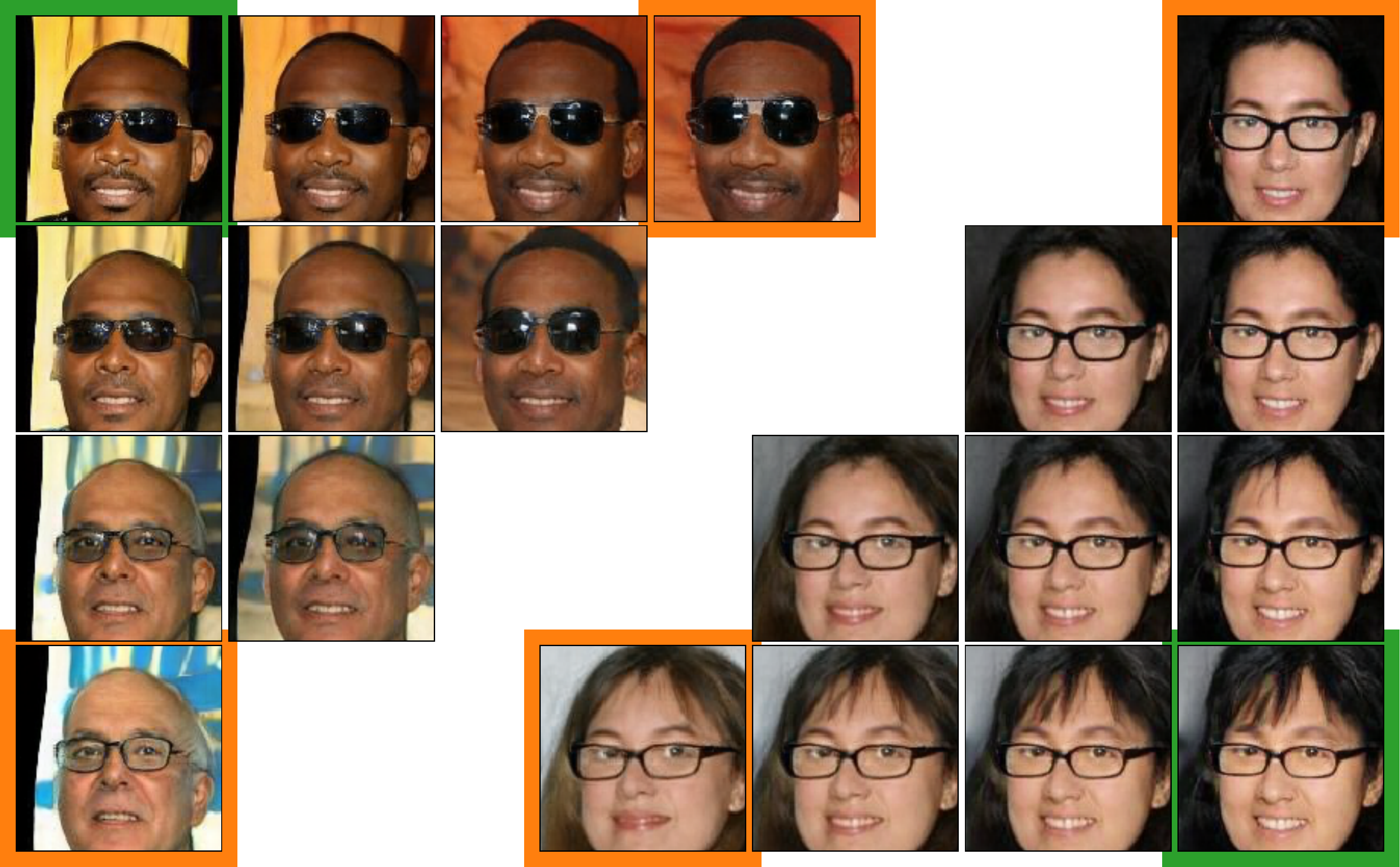}
    \caption{
    Example g-SMOTE augmentations where k=3. 
    Given a datapoint (green) and two neighbours (orange), linear interpolations in GAN latent space yield diverse images. Nearest neighbours are chosen to share the target attribute (\textit{``eyeglasses''}). We give all interpolated images the same attribute label value. 
    }
    \label{fig:g-smote_examples}
\end{figure}

\begin{figure*}
    \centering
    \includegraphics[width=\textwidth]{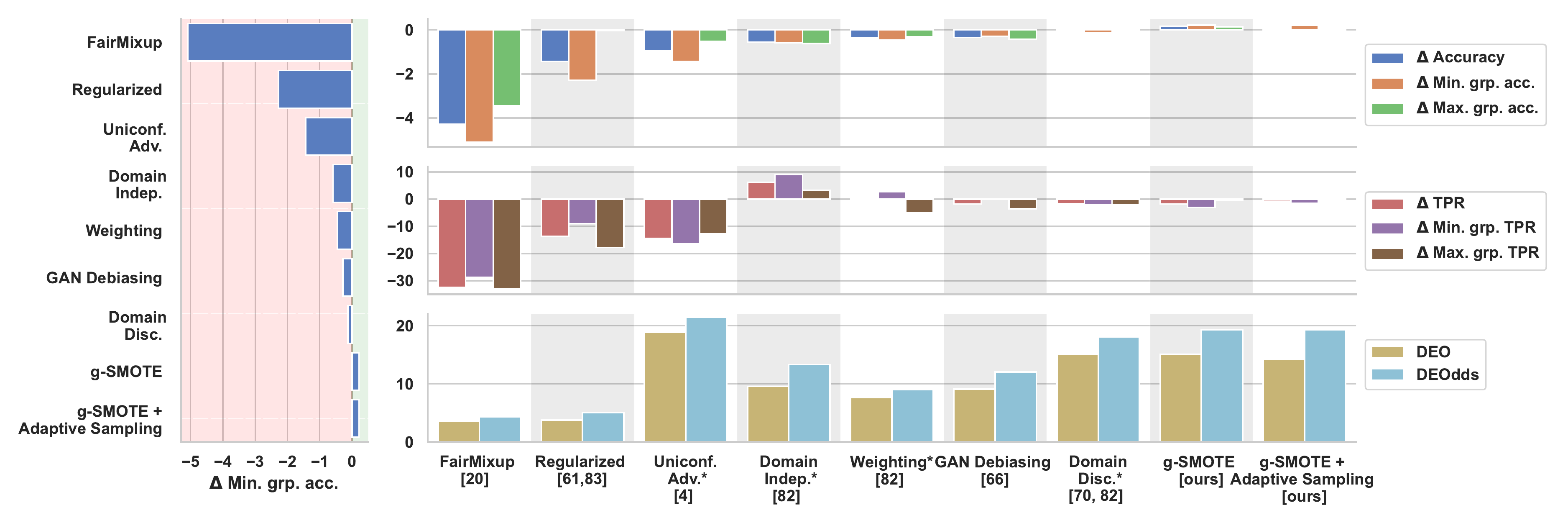}
    \caption{
    Fairness methods on the CelebA dataset. We report mean scores over 13 labels \cite{ramaswamy2020gandebiasing} call \textit{gender independent}. {\bf Left:} Change in minimum group accuracy difference relative to the unconstrained classifier. Only our approaches g-SMOTE and Adaptive g-SMOTE  improve accuracy on the worst performing groups. All other methods lower the accuracy. 
    {\bf Top Right}: A decomposition of the change in accuracy into overall change, change in the best performing group, and the worst.
    {\bf Center Right:} Decomposition of the change in True Positive Rate (TPR). Despite DEO being defined in terms of TPR, all but one of the fairness method decrease the TPR. Note that our g-SMOTE based approaches improve accuracy at the cost of decreasing TPR.
    {\bf Bottom Right:}  plots of DEO and DEOdds. These standard fairness measures  track each other closely, with a high score in one corresponding to a high-score in the other.  However, they are unpredictive of improvements in TPR or accuracy on any groups.
    Methods marked with * are based on the code-base of \cite{wang2020fair}, training one model for multi-task classification. Other methods implement one-task classification models. 
    }
    \label{fig:performance_overview}
\end{figure*}

\begin{figure}
    \centering
    \includegraphics[width=\linewidth]{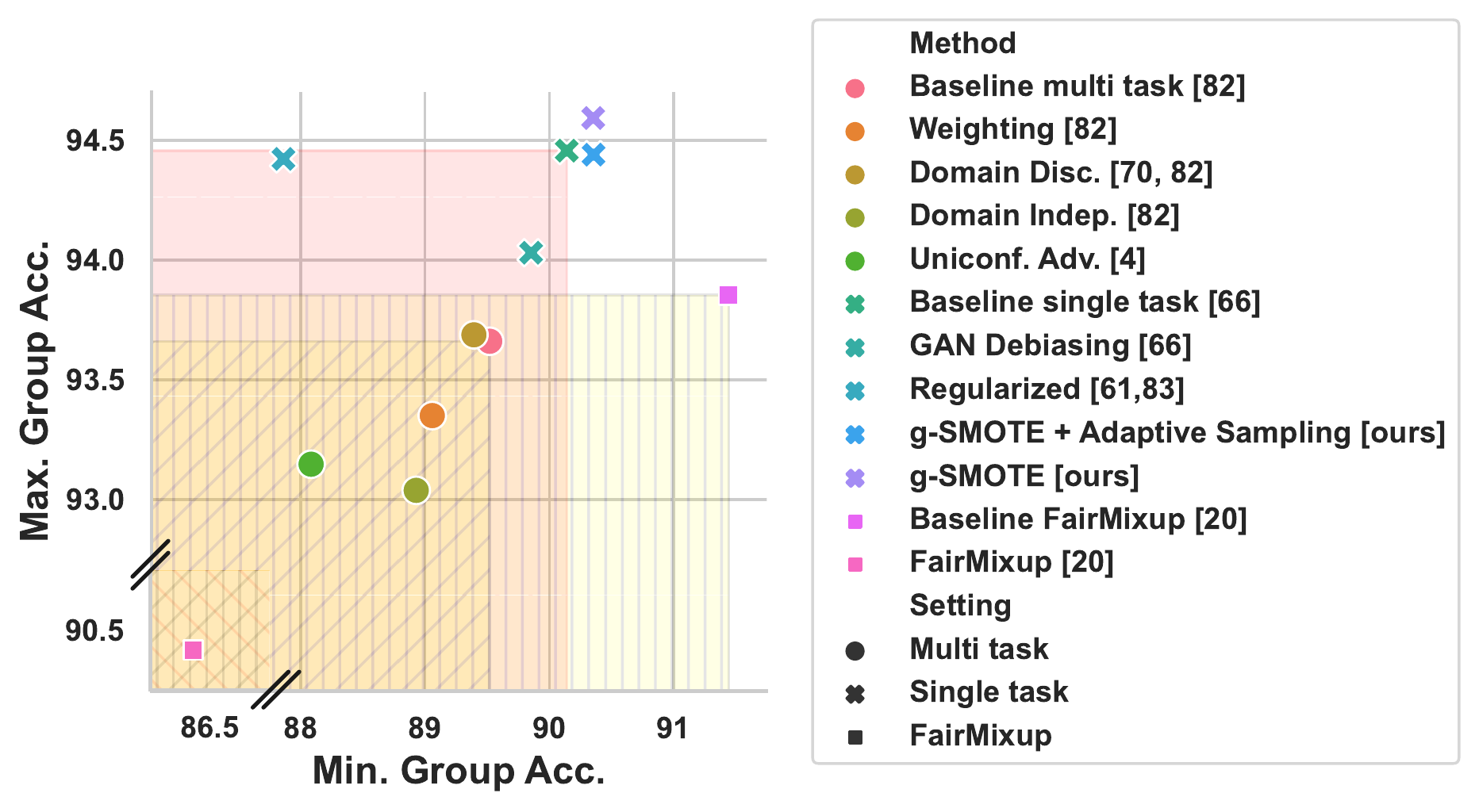}
    \caption{A comparison of fairness methods and unmodified classifiers showing the trade-off behaviour between better and worse performing groups. Results are averaged over $13$ labels with \textit{gender independent quality}. The shaded rectangles indicate Pareto-inefficient areas in which neither of the groups gained performance over the unmodified model. Note that FairMixup uses a smaller model and higher resolution. Our methods are the only ones to improve worst group performance over the relevant baseline.}
    \label{fig:pareto_exp}
\end{figure}
\section{Experiments}\label{sec:experiments}
We show that a wide range of methods for fairness in computer vision decrease performance on the worst group despite reporting improved fairness, and that, in contrast, our approach improves performance. As such, we follow existing experimental protocols as closely as possible. We show  results on CelebA where all methods are reported to work, we use existing code, and compare each method to its corresponding unfair baseline.
We  demonstrate that:
\paragraph{Common fairness approaches in computer vision reduce performance for protected groups compared to unfair baselines.} 
    As such, we evaluate methods with respect to their min-max performance on accuracy and TPR.
\vspace{-0.6em}    
    \paragraph{Adaptive debiasing with g-SMOTE improves min-max performance.} 
    See Section~\ref{sec:exp-minmax}, and Figure~\ref{fig:performance_overview} for experimental results. The benefit of adaptive debiasing is emphasized when training on a subset of data, because 
    the GAN allows for effective unsupervised data augmentation.
\vspace{-0.6em}
    \paragraph{Adaptive debiasing with g-SMOTE works with cross-sectional groups of multiple protected attributes.} 
    
    See Section~\ref{sec:exp-multiple} for experimental results. We show our method works with the four groups defined by the cross product of gender and target attribute binary groups.
\vspace{-0.6em}
    \paragraph{g-SMOTE produces better data diversity than popular augmentation strategies.} 
    See Section~\ref{sec:exp-orth}. When combining g-SMOTE augmentation with popular augmentation methods, such as random flips, crops, rotations and RandAugment, we always observe an improvement in accuracy.

\subsection{Evaluating min-max fairness on existing methods and adaptive debiasing}
\label{sec:exp-minmax}
Our experimental setup follows  \cite{ramaswamy2020gandebiasing} and \cite{wang2020fair} with full details  in Appendix~\ref{sec:architecture}.
We compare against methods to improve fairness, including oversampling, domain discriminative training \cite{saerensAdjusting2002, Royer2015ClassifierAA}, domain independent models \cite{wang2020fair}, an adversarial approach \cite{alvi2019a}, regularization \cite{padala2020fnnc, wick2019}, FairMixup \cite{chuang2021fair} and GAN-based offline dataset debiasing \cite{ramaswamy2020gandebiasing}. 

We summarize results in Figure \ref{fig:performance_overview}. Some methods use a single multitask model (annotated with $*$), others use multiple single task models. Note that FairMixup uses a higher resolution ($256\,\mathrm{px}$) and a lower capacity model (ResNet18), which leads to a substantially different baseline performance. The only methods to increase the performance on the less accurate groups are g-SMOTE with and without adaptive sampling. The baselines lower accuracy across both protected groups. For the offline debiasing approach, we used our own implementation and the same GAN model as used for the g-SMOTE methods. 

The argument set out in Section~\ref{bias_fairness:sec} suggests that for high capacity models, balancing datasets with adaptive sampling is less important than improving  generalization by generating diverse synthetic data. Empirically, we find this to be the case. Using g-SMOTE without adaptive sampling, we find that performance can be improved for all groups, with only a small decrease in min-group accuracy compared to adaptive sampling, see Table \ref{tab:2_groups_sampling}.
We report the min. group accuracy per attribute in Table \ref{tab:per_attribute}.

\subsection{Adaptive g-SMOTE on multiple  groups}
\label{sec:exp-multiple}
Adaptive g-SMOTE is trivially extendable to multiple protected groups. We compare the adaptive g-SMOTE sampling with the regular g-SMOTE augmentation for two protected groups (\textit{``male''} / not \textit{``male''}) and four protected groups (\textit{``male''} / not \textit{``male''}, target attribute / not target attribute) in Table \ref{tab:2_groups_sampling} and Table \ref{tab:4_groups_sampling} respectively. The $32$ target labels are selected such that each of the four protected groups has at least $0.3\, \%$ of the total number of datapoints.
\begin{table}[ht!]
\adjustbox{max width=\linewidth}{
    \begin{tabular}{lr|ccc}
    \toprule
        \textbf{2 Protected Groups} & & \textbf{Baseline} & \textbf{Adaptive g-SMOTE} & \textbf{g-SMOTE}\\
        \midrule
        \makecell[l]{\bfseries 10k train images \\ ~ \\ ~} 
        & \makecell[r]{\bfseries Acc. \\ \bfseries Min. grp. acc. \\ \bfseries DEO\\ \bfseries DEOdds}
        & \makecell{$89.41$ \\ $85.68 $\\ $\mathbf{23.18}$ \\ $\mathbf{33.63}$} 
        & \makecell{$89.57$ \\ $\mathbf{85.93}$ \\ $25.00$ \\ $35.51$} 
        & \makecell{$\mathbf{89.58}$ \\ $85.91$ \\ $24.65$ \\ $34.64$} \\
    \bottomrule
    \end{tabular}
    }
    \caption{Sampling strategies on CelebA. We report the means over $32$ labels. The protected attribute is \textit{``male''}. The models shown have the highest min. group accuracy over the training period.}
    \label{tab:2_groups_sampling}
\end{table}
\begin{table}[ht!]
\adjustbox{max width=\linewidth}{
    \begin{tabular}{lr|ccc}
    \toprule
        \textbf{4 Protected Groups} & & \textbf{Baseline} & \textbf{Adaptive  g-SMOTE} & \textbf{g-SMOTE}\\
        \midrule
        \makecell[l]{\bfseries 10k train images \\ ~ \\ ~} 
        & \makecell[r]{\bfseries Acc. \\ \bfseries Min. grp. acc. \\ \bfseries DEO  \\ \bfseries DEOdds}
        & \makecell{$\mathbf{85.65}$ \\ $62.16$ \\ $19.41$ \\ $28.22$} 
        & \makecell{$84.91$ \\ $\mathbf{64.36}$ \\ $\mathbf{13.25}$ \\ $\mathbf{19.82}$} 
        & \makecell{$85.16$ \\ $63.54$ \\ $19.12$ \\ $27.75$} \\
        \makecell[l]{\bfseries 160k train images \\ ~ \\ ~} 
        & \makecell[r]{\bfseries Acc. \\ \bfseries Min. grp. acc. \\ \bfseries DEO  \\ \bfseries DEOdds}
        & \makecell{$\mathbf{87.37}$ \\ $61.74$ \\ $21.99$ \\ $30.89$} 
        & \makecell{$ 85.77$ \\ $\mathbf{68.06}$ \\ $\mathbf{12.27}$ \\ $\mathbf{19.29}$} 
        & \makecell{$ 87.27$ \\ $ 61.84$ \\ $ 21.91$ \\ $ 31.24$} \\
    \bottomrule
    \end{tabular}
    }
    \caption{Different sampling strategies on CelebA. Reported are the means over $32$ labels. The protected attributes are \textit{``male''} and the target attribute. The models shown have the highest min. group accuracy over the training period. DEO and DEOdds are only evaluated with respect to the attribute \textit{``male''}.}
    \label{tab:4_groups_sampling}
\end{table}
This comparison leads to three interesting findings: (1) g-SMOTE augmentation combined with adaptive sampling yields a significant improvement in min-max fairness when performed on four groups. (2) The performance improvement is even more pronounced when deployed on the full training set %
(see lower half of Table \ref{tab:4_groups_sampling}). (3) Optimizing min~max fairness with the target attribute as an additional protected attribute substantially improves accuracy-based fairness notions, such as DEO. 
\subsection{Orthogonality to other augmentation methods}
\label{sec:exp-orth}
We combine the  g-SMOTE augmentations with a variety of existing augmentation methods, such as random crop, random rotation, random flip and RandAugment. For each of the existing augmentation methods, the g-SMOTE augmentation improves accuracy. This indicates that the data diversity generated by the GAN yields a generalization gain that is at least partially orthogonal to the otherwise achieved improvements. Results are reported in Table \ref{tab:other_augmentations}.
\begin{table}[ht!]
\adjustbox{max width=\linewidth}{
    \begin{tabular}{r|ccccc}
    \toprule
        \textbf{} & \textbf{No Augment} & \textbf{Rand Crop} & \textbf{Rand Rot.} & \textbf{Rand Flip} & \textbf{RandAugment}\\
        \midrule
        \textbf{Without g-SMOTE}      & $89.15$ & $89.56$ & $89.66$  & $89.78$  & $90.17$ \\
        \textbf{With g-SMOTE} & $\mathbf{89.63}$ & $\mathbf{89.85}$ & $\mathbf{89.75}$  & $\mathbf{89.86}$  & $\mathbf{90.33}$ \\
    \bottomrule
    \end{tabular}
    }
    \caption{Min. group accuracy over common augmentations on CelebA  with and without AdaptiveSMOTE sampling. We report mean scores over the labels \cite{ramaswamy2020gandebiasing} call \textit{gender independent}, using the model with greatest min. group accuracy over the training period.}
    \label{tab:other_augmentations}
    \vspace{-4mm}
\end{table}

\section{Discussion}
The majority of fairness methods examined report improved fairness measures of DEO and DEOdds despite decreasing both accuracy and TPR rates of the most disadvantaged groups. Nonetheless, we have shown that it is possible to improve performance for disadvantaged groups by the targeted generation of synthetic data. For the simple case of these high-capacity models (Table~\ref{tab:2_groups_sampling}), in which protected groups correspond to gender, there is no apparent trade-off, and no need to target particular groups with adaptive sampling. Instead, we can uniformly generate samples and improve performance for everyone. In scenarios where we want high accuracy for groups corresponding to a combination of protected attribute and true label (e.g., males receiving a negative decision) this trade-off still exists. Here, adaptive sampling gives a substantial improvement in min-group accuracy as well as improving the  fairness measures.

Based on our theoretical analysis and experimental findings, we make \textit{two key recommendations} for practitioners: 
\textbf{(1) Evaluate a model using the error of the worst performing group:} If  accuracy-based notions of fairness are appropriate for a given scenario, \ie,  particular groups are being disadvantaged by a high error rate, a better measure 
for fairness than the difference of error rates is the error over the worst performing group.  If the  rate is sufficiently low, the method can be safely deployed, otherwise not.  \\
\textbf{(2) Gather more data for the worst performing groups:} We have shown that the key issue preventing fairness in computer vision is an inability to generalize well. However, outside of  standardized benchmarks with fixed training sets, generalization can be improved by gathering more data. This leads to a direct approach where one iteratively evaluates on held-out data and gathers additional data on the worst performing group. This is equivalent to algorithm~1 of \cite{abernethy2020adaptive}, and under certain constraints
is guaranteed to minimize the worst error over all groups.   Other works \cite{gendershades} have called for more diverse datasets, but approaching this as a min-max problem tells us how to grow the dataset, and why diversity improves performance even when it leads to a  distribution mismatch between  training and evaluation data.
\vspace*{-1.1em}
\paragraph{Limitations:} The analysis set out in Section~\ref{analysis} only holds for the accuracy-based notions of fairness that are satisfied by perfect classifiers. It may be possible to enforce other notions of fairness, such as demographic parity, on maximally accurate classifiers without held-out data. Moreover, a wide range of machine learning scenarios exist for which accuracy-based fairness  is inappropriate \cite{wachter2021fairness}. 
\vspace*{-1.1em}
\paragraph{Conclusion:} The takeaway message from this work should not be that accuracy-based notions of fairness do not work in computer vision, but that if we measure and optimize the right thing, they can be made to work.  Our analysis in Section~\ref{analysis} makes it clear that fairness on unseen data is primarily a problem of generalization. 
As such, three promising directions for improving performance in the most disadvantaged groups lie in Hyperparameter Optimization (HO), Network Architecture Search, and data augmentation, with HO already being used in fairness\cite{islam2021can,valdivia2021fair, perrone2021fairbayesian}. We have shown that data augmentation can  improve performance on the most disadvantaged groups, but the use of these other techniques to improve worst group performance is a promising direction for future work. 
\subsection*{Acknowledgments}\noindent
CR is a member of the Governance of Emerging Technologies (GET) research programme at the Oxford Internet Institute. He is grateful to members of GET and the Trustworthy Auditing for AI project for informed discussion that helped shape this work. 
\newpage
{\small
\bibliographystyle{ieee_fullname}
\bibliography{ref}
}

\newpage
\appendix

\renewcommand\thefigure{S\arabic{figure}}
\renewcommand\thetable{S\arabic{table}}
\renewcommand\theequation{S\arabic{equation}}
\renewcommand{\thesection}{\Alph{section}}
\renewcommand{\thealgorithm}{S\arabic{algorithm}}
\setcounter{figure}{0} 
\setcounter{table}{0} 
\setcounter{equation}{0}

\twocolumn[{%
\begin{center}
\textbf{\large{Supplementary Materials}}\\
\vspace{0.5em}
\end{center}
}]
\begin{table*}[ht!]
\adjustbox{max width=\textwidth}{
    \begin{tabular}{l|ccccc|ccccc|cc}
    \toprule
\textbf{Method} & \makecell{\textbf{Baseline}\\\textbf{multi task }\\ ~\textbf{[82] }} & \makecell{\textbf{Weighting }\\ ~\textbf{[82] }} & \makecell{\textbf{Domain Disc. }\\ ~\textbf{[70, 82] }} & \makecell{\textbf{Domain Indep. }\\ ~\textbf{[82] }} & \makecell{\textbf{Uniconf. Adv. }\\ ~\textbf{[4] }} & \makecell{\textbf{Baseline}\\\textbf{single task }\\ ~\textbf{[66]}} & \makecell{\textbf{GAN Debiasing }\\ ~\textbf{[66]}} & \makecell{\textbf{Regularized }\\ ~\textbf{[61,83]}} & \makecell{\textbf{g-SMOTE +}\\\textbf{Adaptive Sampling }\\ ~\textbf{[ours]}} & \makecell{\textbf{g-SMOTE }\\ ~\textbf{[ours]}} & \makecell{\textbf{Baseline}\\\textbf{FairMixup }\\ ~\textbf{[20]  }} & \makecell{\textbf{FairMixup }\\ ~\textbf{[20]  }}\\\midrule
\textbf{Accuracy} & $\mathbf{91.79}$ & $91.45$ & $91.78$ & $91.24$ & $90.86$ & $92.47$ & $92.12$ & $91.05$ & $92.56$ & $\mathbf{92.64}$ & $\mathbf{92.74}$ & $88.46$\\
\textbf{Max. grp. acc.} & $93.66$ & $93.35$ & $\mathbf{93.69}$ & $93.04$ & $93.15$ & $94.46$ & $94.03$ & $94.42$ & $94.44$ & $\mathbf{94.59}$ & $\mathbf{93.85}$ & $90.42$\\
\textbf{Min. grp. acc.} & $\mathbf{89.52}$ & $89.06$ & $89.39$ & $88.93$ & $88.08$ & $90.14$ & $89.85$ & $87.86$ & $\mathbf{90.36}$ & $90.35$ & $\mathbf{91.44}$ & $86.36$\\\midrule
\textbf{TPR} & $64.51$ & $64.02$ & $62.80$ & $\mathbf{70.74}$ & $50.15$ & $\mathbf{67.90}$ & $66.13$ & $54.20$ & $67.11$ & $66.14$ & $\mathbf{79.13}$ & $46.67$\\
\textbf{Max. grp. TPR} & $72.29$ & $67.41$ & $70.13$ & $\mathbf{75.61}$ & $59.59$ & $73.88$ & $70.36$ & $56.11$ & $\mathbf{74.06}$ & $73.43$ & $\mathbf{80.89}$ & $47.85$\\
\textbf{Min. grp. TPR} & $57.09$ & $59.74$ & $55.06$ & $\mathbf{66.05}$ & $40.72$ & $\mathbf{61.34}$ & $61.25$ & $52.34$ & $59.78$ & $58.32$ & $\mathbf{72.92}$ & $44.27$\\\midrule
\textbf{DEO} & $15.20$ & $\mathbf{7.67}$ & $15.08$ & $9.56$ & $18.87$ & $12.54$ & $9.11$ & $\mathbf{3.77}$ & $14.28$ & $15.11$ & $7.97$ & $\mathbf{3.58}$\\
\textbf{DEODD} & $18.14$ & $\mathbf{9.00}$ & $18.10$ & $13.29$ & $21.48$ & $16.54$ & $12.04$ & $\mathbf{5.06}$ & $19.30$ & $19.32$ & $10.06$ & $\mathbf{4.29}$\\

    \bottomrule
    \end{tabular}
    }
    \caption{
     Fairness methods on the CelebA dataset. We report mean scores over the 13 labels \cite{ramaswamy2020gandebiasing} call \textit{gender independent}.
    We report the model with the greatest min. group accuracy over the training period.}
    \label{tab:performance_overview}
\end{table*}

\section{Additional Experimental Results}
An extensive overview for the results visualized in Figure \ref{fig:performance_overview} are listed in Table \ref{tab:performance_overview}.
\begin{table}[ht!]
\begin{center}
\adjustbox{max width=0.9\linewidth}{
    \begin{tabular}{l|cc}
    \toprule
        \textbf{Attribute Name} & \textbf{Baseline} & \textbf{AdaptiveSMOTE}\\
        \midrule
Wavy Hair & $76.95 \pm 0.32$ & $77.86 \pm 0.11$ \\
Big Lips & $80.49 \pm 0.15$ & $80.30 \pm 0.16$ \\
Eyeglasses & $99.23 \pm 0.01$ & $99.25 \pm 0.02$ \\
Attractive & $78.80 \pm 0.14$ & $79.17 \pm 0.10$ \\
Brown Hair & $81.26 \pm 0.05$ & $81.18 \pm 0.12$ \\
Wearing Necklace & $80.30 \pm 0.01$ & $80.30 \pm 0.01$ \\
High Cheekbones & $85.53 \pm 0.11$ & $85.61 \pm 0.22$ \\
Receding Hairline & $91.08 \pm 0.09$ & $91.49 \pm 0.11$ \\
Wearing Hat & $98.21 \pm 0.02$ & $98.22 \pm 0.08$ \\
Black Hair & $86.40 \pm 0.10$ & $87.19 \pm 0.26$ \\
Gray Hair & $95.28 \pm 0.09$ & $95.39 \pm 0.08$ \\
Pale Skin & $95.26 \pm 0.06$ & $95.43 \pm 0.07$ \\
Smiling & $91.64 \pm 0.17$ & $91.91 \pm 0.09$ \\
Chubby & $89.29 \pm 0.02$ & $89.35 \pm 0.12$ \\
Young & $82.28 \pm 0.10$ & $82.99 \pm 0.24$ \\
Wearing Earrings & $83.46 \pm 0.14$ & $83.67 \pm 0.23$ \\
Big Nose & $70.71 \pm 0.29$ & $71.12 \pm 0.32$ \\
Oval Face & $70.67 \pm 0.10$ & $71.15 \pm 0.03$ \\
Bags Under Eyes & $73.24 \pm 0.37$ & $73.69 \pm 0.19$ \\
Bushy Eyebrows & $87.08 \pm 0.09$ & $87.36 \pm 0.08$ \\
Mouth Slightly Open & $93.46 \pm 0.12$ & $93.58 \pm 0.06$ \\
Rosy Cheeks & $90.87 \pm 0.11$ & $90.97 \pm 0.09$ \\
Arched Eyebrows & $76.67 \pm 0.35$ & $76.83 \pm 0.26$ \\
Blurry & $95.58 \pm 0.05$ & $95.61 \pm 0.06$ \\
Wearing Lipstick & $86.14 \pm 0.17$ & $86.44 \pm 0.14$ \\
Blond Hair & $91.96 \pm 0.05$ & $92.10 \pm 0.13$ \\
Heavy Makeup & $84.13 \pm 0.36$ & $84.13 \pm 0.01$ \\
Pointy Nose & $69.20 \pm 0.16$ & $69.86 \pm 0.11$ \\
Straight Hair & $77.72 \pm 0.42$ & $77.98 \pm 0.12$ \\
Bangs & $94.67 \pm 0.17$ & $94.72 \pm 0.04$ \\
Double Chin & $91.43 \pm 0.04$ & $91.57 \pm 0.17$ \\
Narrow Eyes & $91.97 \pm 0.08$ & $92.32 \pm 0.21$ \\
    \bottomrule
    \end{tabular}
    }
\end{center}
    \caption{Min. group accuracy for individual attributes of CelebA. Reported are the means and standard deviations over $3$ restarts of all attributes with at least $11$ positive and negative datapoints per group. The protected attribute is \textit{``male''}. We evaluate the model with greatest min. group accuracy over the training period. Both methods were trained on $10000$ images from the CelebA training set. 
    }
    \label{tab:per_attribute}
\end{table}
\section{GAN based SMOTE}
\begin{algorithm}
\caption{GAN based synthetic minority oversampling}
\begin{algorithmic}[1]
  \scriptsize
  \STATE \textbf{Inputs:} 
  
  Hyper-parameters $m, k \in \mathbb{N}^+$
  
  Train dataset $D_\text{Train} = \{ (x_0, y_0), (x_1, y_1), \dots \}$, $x_i \in \mathcal{X}$, $y_i \in \mathcal{Y}$
  
  Trained invertible GAN $E: \mathcal{X} \rightarrow \mathcal{Z}$, $G: \mathcal{Z} \rightarrow \mathcal{X}, Z = \mathbb{R}^n$
  
  datapoint $(x_s, y_s)$
  
  \STATE \textbf{Encode dataset}
  
  $D_\text{Train} \leftarrow \{ (x_0, y_0, z_0), (x_1, y_1, z_1), \dots \}$, $z_i = E(x_i)$
  
  \STATE \textbf{Determine $m$ nearest neighbours}
  
  $N \leftarrow \{z_0^N, z_1^N, \dots, z_m^N\}$
  
  according to $z_i^N = \argmin_{z_j \not \in \{z_0^N, \dots, z_{i-1}^N\}}  \lVert z_s - z_j \rVert ^2\ \text{s.t.}\ y_s = y_j$
  
  \STATE \textbf{Randomly choose $k$ neighbours from that set}
  
  $N \leftarrow \{ x_{\pi(0)}^N, x_{\pi(1)}^N, \dots, x_{\pi(k)}^N\}$
  
  \STATE \textbf{Uniformly sample}
  $z_\text{Augment} \sim \mathcal{U}\left[ \mathrm{Simplex}\left[N\right] \right]$
  
  \STATE \textbf{Return} $(G(z_\text{Augment}), y_s)$

\end{algorithmic}
\label{alg:augmentation}
\end{algorithm}
\begin{figure}[h!]
    \centering
    \includegraphics[width=4cm]{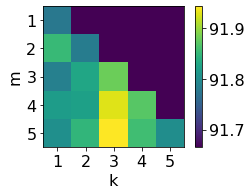}
    \caption{The effect of varying the number of additional neighbors used in the simplex ($k$) against the number of close neighbors sampled from. We see strongest results where $k=3$ (note that $k$ is the number of neighbors, so in this case it corresponds to a $4$-simplex), and where we sample from a greater range of neighbors. $k=1$ corresponds to standard SMOTE.}
    \label{fig:smote_gridsearch}
\end{figure}
\section{Efficient Uniform sampling from the Convex Hull of Points}
\begin{figure}[h!]
    \centering
    \includegraphics[width=0.8\linewidth]{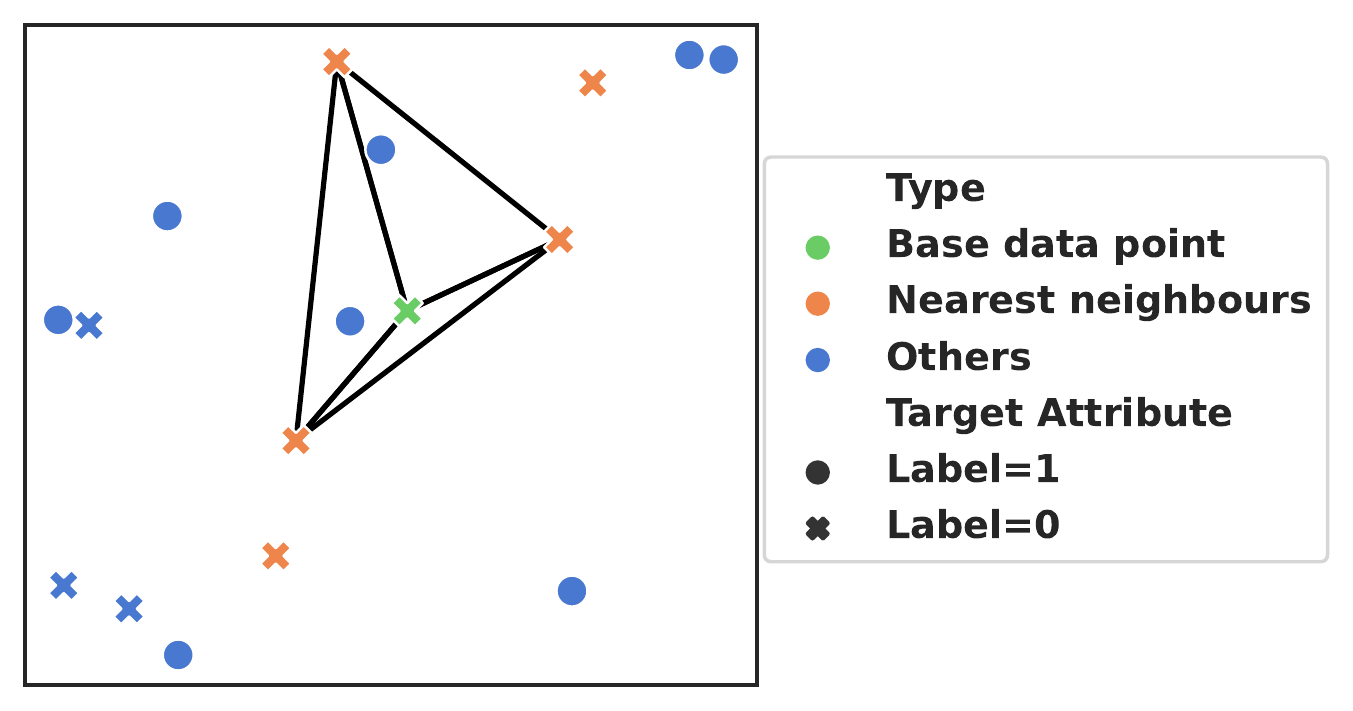}
    \caption{Visualization of the uniform simplex sampling strategy for improved data diversity. Given a datapoint (green) and its nearest $m$ neighbours with the same target label (orange) we randomly choose $k$ neighbours to form a simplex. datapoints with other labels (blue crosses) or too large distance (blue dots) are ignored.}
    \label{fig:smote_simplex_sampling}
\end{figure}
\label{sec:eff_sample}
Here, we briefly summarize the strategy for efficiently uniformly sampling from the convex hull of a set of $k$ points in an $n$-dimensional space where $k\leq n+1$, as we have only been able to find related approaches in the literature for $k=3$ \cite{osada2002shape}.

Under minimal assumptions, e.g., the points are sampled with continuous noise, with probability $1$ the $k$ points form a $k-1$ dimensional simplex, or generalized triangle. 
This allows us to sample efficiently using an inductive process. 
Given a point $\rho_i$ sampled from a $i$ dimensional simplex, $S_i$ we can uniformly sample from an $i+1$ dimensional simplex, $S_{i+1}=S_i \cup \{p_i\}$ by choosing 
\begin{equation}
    \rho_{i+1}=\lambda^{1/i} \rho_i + (1-\lambda^{1/i})p_{i+1}
\end{equation}
where $\lambda$ is sampled from the uniform distribution $U[0,1]$.

The base case is $S_1=\{p_1\}$, and $\rho_1=p_1$.

Note that when the assumption fails, and points are colinear and do not form a simplex, the algorithm degrades reasonably. Sampled points will lie inside the convex hull of points, but the samples will not be uniform.  

\section{Architecture, Dataset and Details}
\label{sec:architecture}
\looseness-1The classifiers are built upon pretrained ResNet50 models \cite{he2016deep}. 
For each of the binary attributes of the CelebA dataset, one classifier is trained. 
As the protected attribute we chose \textit{``male''}\footnote{The labels in CelebA refer to an externally assigned perceived binary gender, and not to self-assigned gender identity. 
Although the binary nature of the label does not reflect the true distribution of either, we are restricted to the annotations available in the dataset.}.
Each model was trained for $3\cdot 10^6$ images and evaluated every $500$ batches using Adam ($lr=10^{-4}$) and a batch size of $64$. 
Images were center-cropped and down-scaled to $128 \times 128$. 
We use RandAugment with $N=3$, $M=15$ for every experiment unless otherwise stated. 
The reported numbers are the retrieved peak performance during the training period, evaluated on the hold-out evaluation dataset. 
We rely on the analysis of \cite{ramaswamy2020gandebiasing} and report the means over the attributes with \textit{gender independent label quality}: ``bags under eyes'', ``bangs'', ``black hair'', ``blond hair'', ``brown hair'', ``chubby'', ``eyeglasses'', ``gray hair'', ``high cheekbones'', ``mouth slightly open'', ``narrow eyes'', ``smiling'', ``wearing hat''. 
Rows labelled with * show results achieved using the codebase of \cite{wang2020fair}.
For the invertible GAN model, we choose InvGAN \cite{ghosh2021invgan}, however the requirements on the model are low: Aside from state-of-the-art image quality, we need decent interpolation capabilities as well as a reasonable semantic structure of the latent space.

\paragraph{Regularized Approach.} The regularized model presented in Figure~\ref{fig:performance_overview} is based on a regularizer used in \cite{wick2019}. Given training data $\mathcal{D} = \{x_i, y_i, a_i\}_{i=1}^n$ and a model $f$ (as above) that outputs classification scores in $[0,1]$, we define the regularizer~$\mathcal{R}^{\mathrm{DEO}}$ as
\begin{align*}
     \mathcal{R}^{\mathrm{DEO}}(f) := \left(\frac{1}{n_{11}} \sum\limits_{\substack{ \mathcal{D}_{11} }} f(x_i)- \frac{1}{n_{10}} \sum\limits_{\substack{ \mathcal{D}_{10} }} f(x_i)\right)^2, 
\end{align*}
where $\mathcal{D}_{ya} = \{(\tilde{x}, \tilde{y}, \tilde{a}): \tilde{a}=a, \tilde{y} = y, (\tilde{x}, \tilde{y}, \tilde{a}) \in \mathcal{D}\}$ and $n_{ya} = |\mathcal{D}_{ya}|$ for $g \in \{0, 1\}$.

We simply add the fairness regularizer to the loss and trade-off fairness with the accuracy of 
the classifier via a hyperparameter~$\lambda$. We minimize  $\widehat{\mathrm{L}}(f) = \sum_{i=1}^n l(f(x_i),y_i) + \lambda   \mathcal{R}^{\mathrm{DEO}}(f)$.
\end{document}